\documentclass[11pt,a4paper]{article}
\usepackage[hyperref]{emnlp2020}
\usepackage{times}
\usepackage{latexsym}

\usepackage{times}
\usepackage{url}
\usepackage{latexsym}
\usepackage{graphicx}
\usepackage{amsmath}
\usepackage{amsthm}
\usepackage{booktabs}       
\usepackage{amsmath}
\usepackage{enumerate}
\usepackage{amsfonts}
\usepackage{algorithm}
\usepackage{algorithmic}
\usepackage{nicefrac}       
\usepackage{microtype}      
\usepackage{subfigure}
\usepackage{graphicx}
\usepackage{float}
\usepackage{bbm}
\usepackage{multirow}

\usepackage{colortbl}
\definecolor{gray}{RGB}{87, 87, 87}
\definecolor{red}{RGB}{173, 35, 35}
\definecolor{blue}{RGB}{42, 75, 215}
\definecolor{green}{RGB}{29, 105, 20}
\definecolor{brown}{RGB}{129, 74, 25}
\definecolor{purple}{RGB}{129, 38, 192}
\definecolor{cyan}{RGB}{41, 208, 208}
\definecolor{yellow}{RGB}{189, 167, 0}
\definecolor{Red}{rgb}{0.68, 0.05, 0.0}
\definecolor{Blue}{rgb}{0.0, 0.0, 0.61}
\definecolor{Blue1}{RGB}{214, 235, 245}
\definecolor{Blue2}{RGB}{235, 245, 250}
\definecolor{lime}{RGB}{60,179,113}
\definecolor{peach}{RGB}{255, 242, 230}
\usepackage{microtype}

\aclfinalcopy %

\title{Diverse, Controllable, and Keyphrase-Aware:\\ A Corpus and Method for News Multi-Headline Generation}

\author{Dayiheng Liu$^\spadesuit$\thanks{\hspace{2mm}Work is done during internship at Microsoft Research Asia.}\:\:\:\: Yeyun Gong$^{\dag}$ \:\:\:\: Yu Yan$^{\ddag}$ \:\:\:\: Jie Fu$^\diamondsuit$ \\ 
\textbf{Bo Shao$^{\dag}$ \:\:\:\: Daxin Jiang$^{\ddag}$  \:\:\:\: Jiancheng Lv$^{\spadesuit}$   \:\:\:\: Nan Duan$^{\dag}$  }   \\
$^\spadesuit$College of Computer Science, Sichuan University    \:\:\:\:  $^\dag$Microsoft Research Asia  \\  $^\diamondsuit$ Mila \:\:\:\:  $^\ddag$Microsoft \\
losinuris@gmail.com \:\:\:\: 
}

\date{}

\begin{document}
\maketitle
\begin{abstract}
News headline generation aims to produce a short sentence to attract readers to read the news. One news article often contains multiple keyphrases that are of interest to different users, which can naturally have multiple reasonable headlines. However, most existing methods focus on the single headline generation. In this paper, we propose generating multiple headlines with keyphrases of user interests, whose main idea is to generate multiple keyphrases of interest to users for the news first, and then generate multiple keyphrase-relevant headlines. We propose a multi-source Transformer decoder, which takes three sources as inputs: (a) keyphrase, (b) keyphrase-filtered article, and (c) original article to generate keyphrase-relevant, high-quality, and diverse headlines. Furthermore, we propose a simple and effective method to mine the keyphrases of interest in the news article and build a first large-scale keyphrase-aware news headline corpus, which contains over 180K aligned triples of $\langle$news article, headline, keyphrase$\rangle$. Extensive experimental comparisons on the real-world dataset show that the proposed method achieves state-of-the-art results in terms of quality and diversity\footnote{The source code will be available at \url{https://github.com/dayihengliu/KeyMultiHeadline}.}.
\end{abstract}

\section{Introduction} \label{sec:intro}
News Headline Generation is an under-explored subtask of text summarization~\cite{see2017get,gehrmann2018bottom,zhong2019searching}. 
Unlike text summaries that contain multiple context-related sentences to cover the main ideas of a document, news headlines often contain a single short sentence to encourage users to read the news.
Since one news article typically contains multiple keyphrases or topics of interest to different users, it is useful to generate multiple headlines covering different keyphrases for the news article.
Multi-headline generation aims to generate multiple independent headlines, which allows us to recommend news with different news headlines based on the interests of users. 
Besides, multi-headline generation can provide multiple hints for human news editors to assist them in writing news headlines.

However, most existing methods~\cite{takase2016neural,ayana2016neural,murao2019case,colmenares2019headline,zhang2018question} focus on single-headline generation.
The headline generation process is treated as an one-to-one mapping (the input is an article and the output is a headline), which trains and tests the models without any additional guiding information or constraints. 
We argue that this may lead to two problems. 
Firstly, since it is reasonable to generate multiple headlines for the news, training to generate the single ground-truth might result in a lack of more detailed guidance. 
Even worse, a single ground-truth without any constraint or guidance is often not enough to measure the quality of the generated headline for model testing.
For example, even if a generated headline is considered reasonable by humans, it can get a low score in ROUGE~\cite{lin2004rouge}, because it might focus on the keyphrases or aspects that are not consistent with the ground-truth. 

In this paper, we incorporate the keyphrase information into the headline generation as additional guidance. 
Unlike one-to-one mapping employed in previous works, we treat the headline generation process as a two-to-one mapping, where the inputs are news articles and keyphrases, and the output is a headline.
We propose a keyphrase-aware news multi-headline generation method, which contains two modules: (a) Keyphrase Generation Model, which aims to generate multiple keyphrases of interest to users for the news article. (b) Keyphrase-Aware Multi-Headline Generation Model, which takes the news article and a keyphrase as input and generates a keyphrase-relevant news headline.
For training models, we build a first large-scale news keyphrase-aware headline corpus that contains 180K aligned triples of $\langle$news article, headline, keyphrase$\rangle$ .

\begin{figure}[th] 
  \centering
  \includegraphics[scale=0.43]{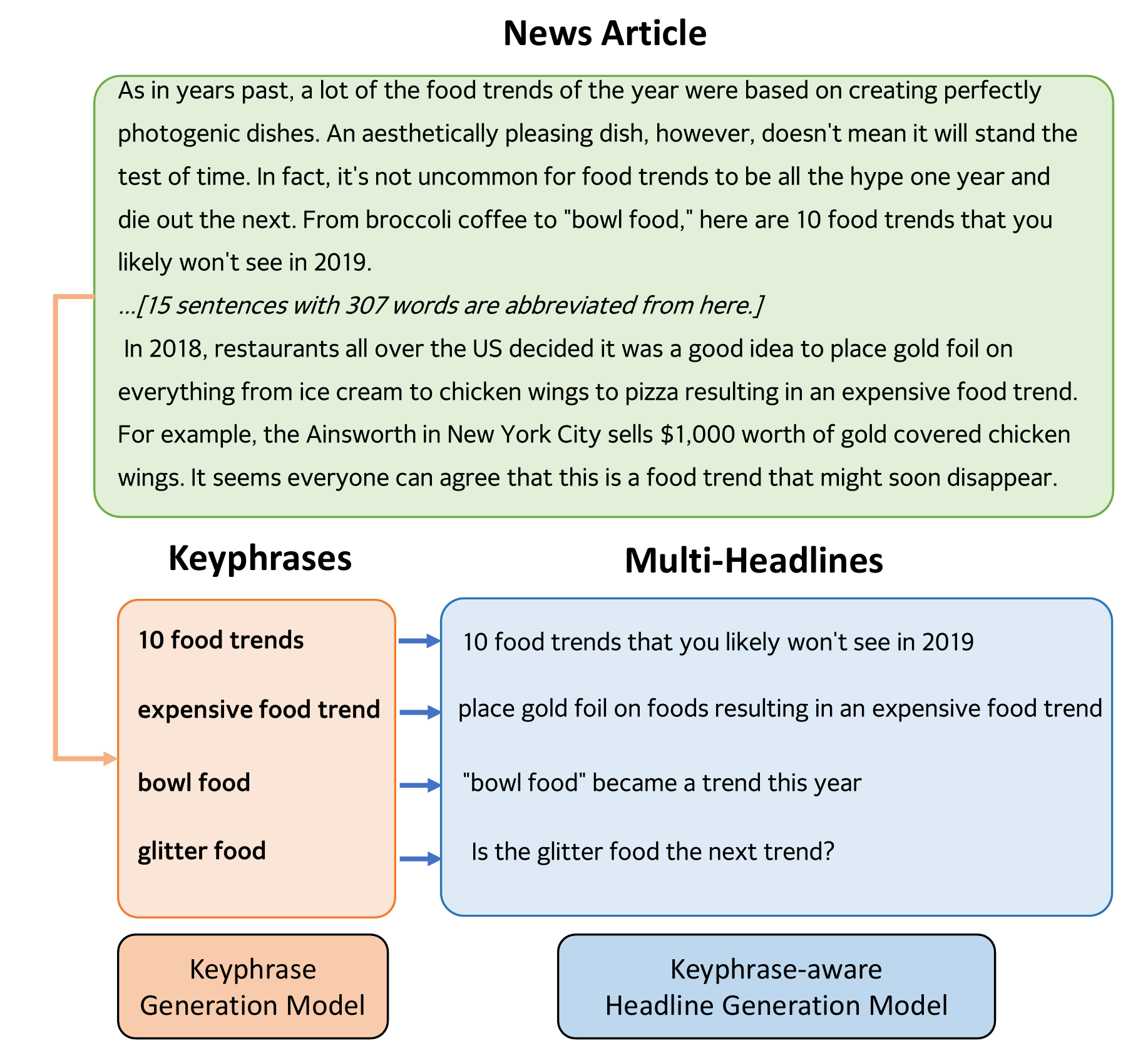}
  \caption{Keyphrase-aware multi-headline generation} \label{fig:frame}
\end{figure}

The proposed approach faces two major challenges. 
The first one is how to build the keyphrase-aware news headline corpus. 
To our best knowledge, no corpus contains the news article and headline pairs, which are aligned with a keyphrase of interest to users. 
The second is how to design the keyphrase-aware news headline generation model to ensure that the generated headlines are keyphrase-relevant, high-quality, and diverse.
For the first challenge, we propose a simple but efficient method to mine the keyphrases of interest to users in news articles based on the user search queries and news click information that are collected from a real-world search engine.
With this method, we build the keyphrase-aware news headline corpus. 

For the second challenge, we design a multi-source Transformer~\cite{vaswani2017attention} decoder to improve the generation quality and the keyphrase sensitivity of the model, which takes three source information as inputs: (a) keyphrase, (b) keyphrase-filtered article, and (c) original article. 
For the proposed multi-source Transformer decoder, we further design and compare several variants of attention-based fusing mechanism. 
Extensive experiments on real-world dataset have shown that the proposed method can generate high-quality, keyphrase-relevant, and diverse news headlines.

\section{Keyphrase-Aware Headline Corpus}
Our keyphrase-aware news headline corpus called \textbf{KeyAware News} is built by the following steps:

(1) \textbf{Data Collection}. We collect 16,000,000 raw samples which contain news articles with user search query information from Microsoft Bing News search engine\footnote{All news articles are high-quality, real-world news and all the news headlines are written by humans.}. Each sample can be presented as a tuple $\langle Q, X, Y, C \rangle$ where $Q$ is a user search query, $X$ is a news article that the search engine returns to the user based on the search query $Q$, $Y$ is a human-written headline for $X$, and $C$ represents the number of times the user clicks on the news under the search query $Q$. Each news article $X$ has 10 different queries $Q$ on average.

(2) \textbf{Keyphrase Mining}. We mine the keyphrase of interest to users with user search queries.
We assume that if many users find and click on one news article through different queries containing the same phrase, such a phrase is the keyphrase for the article.
For each article, we collect its corresponding user search queries and remove the stop words and special symbols from the queries.
Then we find the common phrases (4-gram, 3-gram, or 2-gram) in these queries. 
These common phrases are scored based on how many times they appear in these queries and normalized by length. The score is also weighted by the user click number $C$, which means the phrases that appear in the queries have more users click on the article are more important.
Finally, we use the $n$-gram with the highest score as the keyphrase $Z$ of the article $X$.

(3) \textbf{Article-Headline-Keyphrase Alignment}. In order to obtain the aligned article-headline-keyphrase tuple $\langle X, Y, Z \rangle$. We filter out the sample whose article or headline does not contain the $Z$. Moreover, we remove such pairs whose article length are greater than 600 or less than 100 tokens, or whose headline length are greater than 20 or less than 3 tokens. After the alignment and data cleaning, we obtain the \textbf{KeyAware News} which contains about 180K aligned article-headline-keyphrase triples. We split it into Train, Test, and Dev sets, each containing 165,913, 10,000, and 5,000 samples. 
\begin{figure*}[t] 
  \centering
  \includegraphics[scale=0.3]{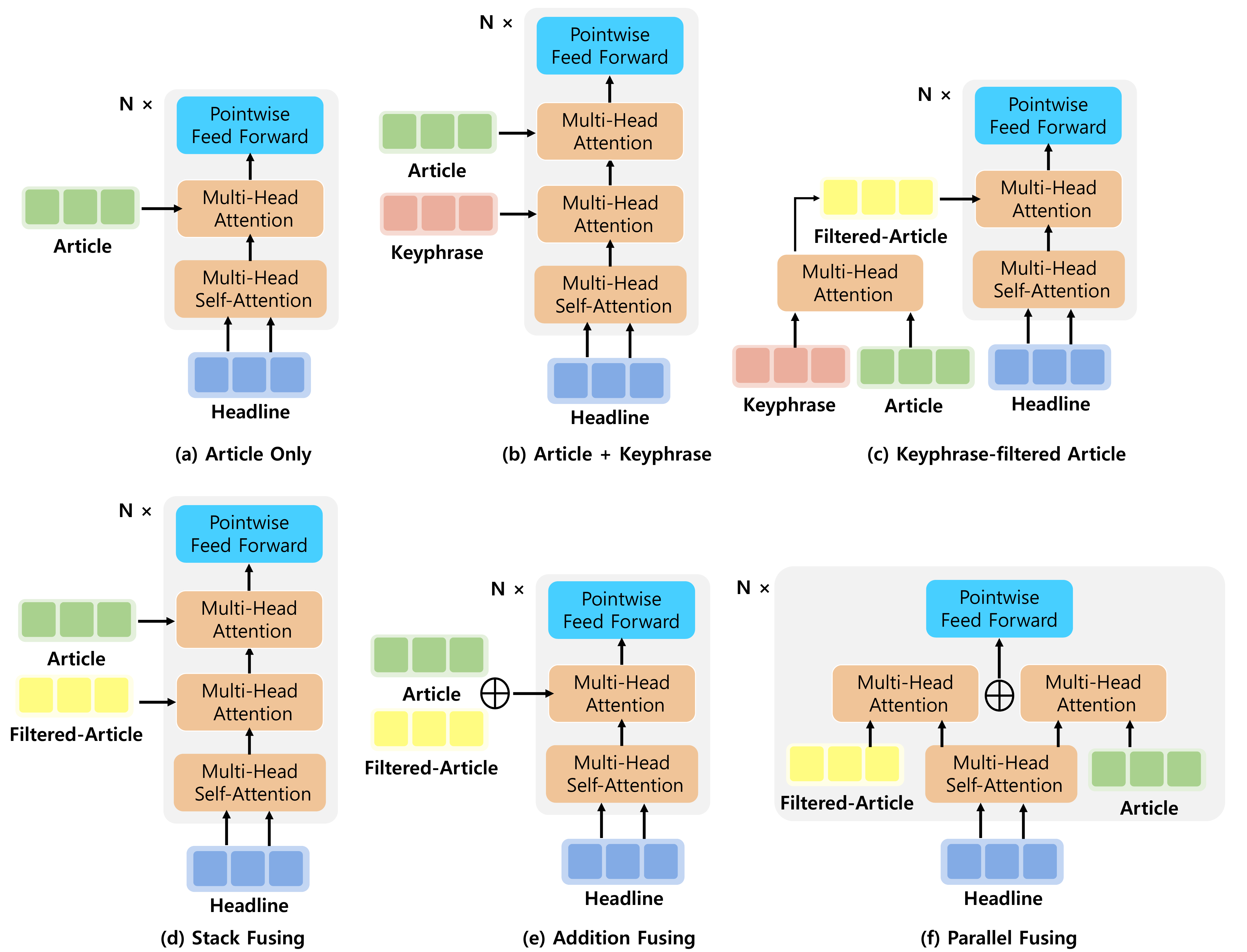}
  \caption{Visualization of the computational steps in each block of our multi-source Transformer decoders.} \label{fig:models}
\end{figure*}

\section{Methodology}
\subsection{Overview}
The overall keyphrase-aware multi-headline generation procedure is shown in Figure~\ref{fig:frame}, which involves two modules: (a) \textbf{keyphrase generation model} generates multiple keyphrases of interest to users for the news article. (b) \textbf{keyphrase-aware headline generation model} takes the news article and each generated keyphrase as input, and generates multiple keyphrase-relevant news headlines.

\subsection{Headline Generation}
The headline generation can be formalized as a sequence-to-sequence learning~\cite{sutskever2014sequence} task.
Given an input news article $X$ and a specific keyphrase $Z$, we aim to produce a keyphrase-relevant headline $Y$.

\subsubsection{Headline Generation BASE Model} \label{sec:model}
We first introduce the basic version of our headline generation model (we call BASE), which is keyphrase-agnostic. BASE is built upon the Transformer Seq2Seq model~\cite{vaswani2017attention}, which has made remarkable progress in sequence-to-sequence learning. Transformer contains a multi-head self-attention encoder and a multi-head self-attention decoder. 
As discussed in \citet{vaswani2017attention}, an attention function maps a query and a set of key-value pairs to an output as: 
\begin{equation}
\textbf{Attention}(\bar{Q}, \bar{K}, \bar{V}) = \textbf{Softmax}(\frac{\bar{Q} \bar{K}^T}{\sqrt{d_k}})\bar{V}, \notag
\end{equation}
where the queries $\bar{Q}$, keys $\bar{K}$, and values $\bar{V}$ are all vectors, and $d_k$ is the dimension of the key vector.
Multi-head attention mechanism projects queries, keys, and values to $h$ different subspaces and calculates corresponding attention as:
\begin{align}
 &\textbf{MultiHead}(\bar{Q}, \bar{K}, \bar{V}) = \textbf{Concat}(\rm{h_1,...,h_h})W^O, \notag \\
 &\rm{where} \; h_i = \textbf{Attention}(\bar{Q}W^Q_i, \bar{K}W^K_i, \bar{V}W^V_i). \notag
\end{align}
The encoder is composed of a stack of $N$ identical blocks. Each block has two sub-layers: multi-head self-attention mechanism and a position-wise fully connected feed-forward network. All sub-layers are interconnected with residual connections~\cite{he2016deep} and layer normalization~\cite{ba2016layer}.

Similarly, the decoder is also composed of a stack of $N$ identical block. In addition to the two sub-layers in each encoder block, the decoder contains a third sub-layer which performs multi-head attention over the output of the encoder.
Figure~\ref{fig:models} (a) shows the architecture of the block in the decoder.

BASE uses the pre-trained BERT-base model~\cite{devlin2018bert} to initialize the parameters of the encoder. 
Also, it uses the transformer decoder with a copy mechanism~\cite{gu2016incorporating}, whose hidden size, the number of multi-head $h$, and the number of blocks $N$ are the same as its encoder.

\subsubsection{Keyphrase-Aware Headline Generation Model}
In order to explore more effective ways of incorporating keyphrase information into BASE, we design 5 variants of multi-source Transformer decoders.

\noindent \textbf{Article + Keyphrase}.
The basic idea is to add the keyphrase into the decoder directly. 
The keyphrase $X_{key}$ is represented as a sequence of word embeddings. 
As shown in Figure~\ref{fig:models} (b), we add an extra sub-layer that performs multi-head attention over the $X_{key}$ in each block of the decoder.
\begin{equation}
X_{dec}^{(n+1)} = \textbf{MultiHead}(X_{dec}^{(n)}, X_{key}, X_{key}),
\end{equation}
where $X_{dec}^{(n)}$ is the output of the $n$-th block in the decoder.
Since the original article has contained sufficient information for the model to learn to generate the headline, the model may tend to mainly use the article information and ignore the keyphrase and become less sensitive to keyphrases. 
As a byproduct, the generated headlines may lack diversity and keyphrase relevance.

\noindent \textbf{Keyphrase-Filtered Article}.
Intuitively, when people read news articles, they tend to focus on the parts of the article that are matched to the keyphrases of their interests. 
Inspired by this, before inputting the original article representation into the decoder, we use the attention mechanism to filter the article with the keyphrase (see Figure~\ref{fig:models} (c)). 
\begin{equation}
\hat{X}_{enc} = \textbf{MultiHead}(X_{key}, X_{enc}, X_{enc}),
\end{equation}
where $X_{enc}$ is the output of the last block in the encoder.
The resulting representation $\hat{X}_{enc}$ can be seen as the keyphrase-filtered article, which mainly keeps the article information that is related to the keyphrase.
Since the decoder cannot directly access the representation of the original article, the model is forced to utilize the information of the keyphrase. Therefore, the sensitivity of the model to keyphrase is improved. 

\noindent \textbf{Fusing Keyphrase-Filtered Article and Original Article}. 
Although feeding the keyphrase-filtered article representation $\hat{X}_{enc}$ instead of the original article representation $X_{enc}$ to the decoder can improve the sensitivity of the model to keyphrase, some useful and global information in the original article may also be filtered out.
It might reduce the quality of the generated headlines.
To further balance the keyphrase sensitivity and headline quality of the model, we use $X_{enc}$ and $\hat{X}_{enc}$ as two input sources for the decoder and fuse them.
As shown in Figure~\ref{fig:models} (d)-(f), we design three decoder variants based on different fusing mechanism to fuse the $X_{enc}$ and the $\hat{X}_{enc}$.

\noindent (I) \textbf{Addition-Fusing Mechanism}. We directly perform a point-wise addition between the $X_{enc}$ and the $\hat{X}_{enc}$. Then we feed it into the decoder.

\noindent (II) \textbf{Stack-Fusing Mechanism}. We perform a multi-head attention on $\hat{X}_{enc}$ and $X_{enc}$ one by one in each block of the decoder. All of the sub-layers are interconnected with residual connections. 
\begin{align}
    \hat{X}_{dec}^{(n)} &= \textbf{MultiHead}(X_{dec}^{(n)}, \hat{X}_{enc}, \hat{X}_{enc}) \\ 
    X_{dec}^{(n+1)} &= \textbf{MultiHead}(\hat{X}_{dec}^{(n)}, X_{enc}, X_{enc})
\end{align}

\noindent (III) \textbf{Parallel-Fusing Mechanism}. For each block of the decoder, we perform a multi-head attention in parallel on $\hat{X}_{enc}$ and $X_{enc}$. Then, we perform a point-wise addition between them. Similarly, all of the sub-layers are interconnected with residual connections. 
\begin{align}
    \hat{X}_{dec}^{(n)} &= \textbf{MultiHead}(X_{dec}^{(n)}, \hat{X}_{enc}, \hat{X}_{enc}) \\ 
    \bar{X}_{dec}^{(n)} &= \textbf{MultiHead}(X_{dec}^{(n)}, X_{enc}, X_{enc}) \\
    X_{dec}^{(n+1)} &= \bar{X}_{dec}^{(n)} + \hat{X}_{dec}^{(n)}
\end{align}

\subsection{Keyphrase Generation}
In this subsection, we show how to generate the keyphrases for a given news article $X$. Here we briefly describe three methods for keyphrase generation. 
It should be noted that in this paper, we mainly focus on news headline generation rather than keyphrase generation.

\noindent (1) \textbf{TF-IDF Ranking}. We use Term Frequency Inverse Document Frequency (TF-IDF) \cite{zhang2007comparative} to weight all $n$-grams ($n=2$, 3, and 4) in the news article $X$. Then we filter out $n$-grams with TF-IDF below the threshold or containing any punctuation or special character. For different $n$ of the $n$-gram, we set different thresholds for filtering. We take this unsupervised method as a baseline.

\noindent (2) \textbf{Seq2Seq}. Since our \textbf{KeyAware News} corpus contains the article-keyphrase pairs, we treat the keyphrase generation as a sequence-to-sequence learning task. 
We train the model BASE with article-keyphrase pairs. During inference, we use beam search with length penalty to generate $n$-grams ($n=2$, 3, and 4) as the keyphrases. 

\noindent (3) \textbf{Slot Tagging}. Because the keyphrases also appear in the news articles, we can formulate the keyphrase generation task as a slot tagging task~\cite{zhang2016keyphrase,williams2019neural}.
We fine-tune the BERT-base model to achieve that.
Concretely, we use the output sequence of the model to predict the beginning and end position of the keyphrase in the article.
During inference, we follow the answer span prediction method used in~\citet{seo2016bidirectional} to predict $n$-grams ($n=2$, 3, and 4) with the highest probabilities as the keyphrases.

\section{Experiments}

\begin{table}[t]
\centering
\scriptsize
  \begin{tabular}{ccccccccl}
    \toprule
    Method & EM@1 & EM@3 & EM@5 & R@1 & R@3 & R@5\\
    \midrule
    TF-IDF & 18.63 & 42.05 & 52.60 & 30.13 & 53.82 & 63.91 \\
    SEQ2SEQ    & 57.27 & \textbf{78.45} & \textbf{84.26} & 59.60 & 81.32 & 87.04 \\
    SLOT   & \textbf{60.75} & 76.94 & 83.18 & \textbf{65.13} & \textbf{84.05} & \textbf{89.08} \\
 \bottomrule
\end{tabular}
\caption{Keyphrase Generation Results}
  \label{tab:key_gen}
\end{table}

\subsection{Keyphrase Generation}
In the first experiment, we evaluate the performance of three keyphrase generation methods: (a) unsupervised TF-IDF Ranking, (b) supervised sequence-to-sequence model (SEQ2SEQ), and (c) supervised slot tagging model (SLOT). 

\noindent \textbf{Implementation and Hyperparameters}. The SEQ2SEQ has the same architecture hyperparameters as BASE model. 
And the architecture hyperparameters of SLOT are the same as those of the BERT-base\footnote{\url{https://github.com/google-research/bert}.}.
We use article-keyphrase pairs in the train set of \textbf{KeyAware News} to train SEQ2SEQ and SLOT.

\noindent \textbf{Metrics}. For evaluation, each method generates top-$K$ keyphrases for every news article in the test set.
We use a top-$K$ exact-match rate (EM@$K$) as an evaluation metric, which tests whether one of the $K$ generated keyphrases matches the golden keyphrase exactly.
Some of the generated key phrases may not exactly match the golden keyphrase but have overlapping tokens with it (it may be a sub-sequence of the golden keyphrase or vice versa). We thus report the Recall@$K$ (R@$K$), which tests the percentage of the tokens in golden keyphrase covered by the $K$ generated keyphrases.

\noindent \textbf{Results}. The results are shown in Table~\ref{tab:key_gen}.
We can see that the EM@1 of TF-IDF is only 18.63\%, but SLOT achieves 60.75\%. 
Both of SEQ2SEQ and SLOT significantly outperform the TF-IDF in all metrics.
SEQ2SEQ achieves comparable performances in EM@$K$, but performs worse than SLOT in R@$K$.
SLOT achieves 83.18\% EM@5 and 89.08\% R@5.
In the following experiments, we use SLOT to generate keyphrases for our keyphrase-aware news headline generation models.

\subsection{News Headline Generation}

\begin{table*}[t]
\centering
\scriptsize
  \begin{tabular}{|l|c|c|c|c|c|c|c|c|c|c|c|c|c|c|c|c|c|c|c|}
    \hline
    \multirow{2}{*}{Method} & \multicolumn{3}{c|}{ROUGE-1} & \multicolumn{3}{c|}{ROUGE-2} & \multicolumn{3}{c|}{ROUGE-L} & \multicolumn{2}{c|}{Distinct-1} & \multicolumn{2}{c|}{Distinct-2}\\ \cline{2-14} 
    &K=1 & K=3 &K=5 & K=1 &K=3 & K=5 & K=1 & K=3 & K=5 & K=3 & K=5 & K=3 & K=5 \\
    \hline
    PT-GEN & 35.66 & 39.82 & 41.59 & 19.80 & 22.60 & 23.84 & 32.96 & 36.73 & 38.33 & 0.125 & 0.076 & 0.215 & 0.143   \\
    SEASS  & 31.20 & 34.52 & 35.98 & 14.82 & 16.52 & 17.17  & 28.23 & 31.04 & 32.25 & 0.112 & 0.069 & 0.191 & 0.126   \\
    Transformer + Copy 	&38.91 & 43.80 & 45.72 & 21.85 & 25.31 & 26.69 & 35.32 & 39.65 & 41.38 & 0.110 & 0.059 & 0.183 & 0.111  \\
    \hline
    BASE  & 42.09 & 45.40 & 47.21 & 24.10 & 26.70 & 28.13 & 38.36 & 41.33 & 42.98 & 0.131 & 0.074 & 0.223 & 0.139  \\
    BASE + Diverse  & - & 45.83 & 47.89 & - & 26.62 & 28.28 & - & 41.77 & 43.71 & 0.182 & 0.111 & 0.313 & 0.213  \\ \hline
    BASE + Filter   & 39.44 & 41.52 & 43.89 & 20.82 & 21.81 & 23.60 & 35.30 & 37.12 & 39.38 & \textbf{0.378} & \textbf{0.294} & \textbf{0.637} & \textbf{0.575}  \\
    BASE + KEY   &43.53 & 47.07 & 49.08 & 25.27 & 27.81 & 29.44 & 39.50 & 42.76 & 44.67 & 0.193 & 0.121 & 0.309 & 0.218  \\ \hline 
    BASE + AddFuse  & \textbf{44.30} & 47.36 & 49.46 & \textbf{25.98} & 27.39 & 29.11 & \textbf{40.24} & 42.48 & 44.47 & 0.235 & 0.156 & 0.385 & 0.290  \\ 
    BASE + ParallelFuse  &43.74 & 47.28 & 49.69 & 25.20& 27.56 & 29.49 & 39.41 & 42.50 & 44.77 & 0.261 & 0.177 & 0.430 & 0.333  \\
    BASE + StackFuse  & 43.97 & 47.63 & 49.74 & 25.32 & 27.90 & 29.69 & 39.60 & 42.96 & 44.97 & 0.201 & 0.127 & 0.332 & 0.237  \\ \hline 
    BASE + AddFuse + KEY  &43.12 & 46.82 & 49.16 & 24.66 & 27.09 &  28.91 & 38.91 & 42.18 & 44.37 & 0.276 &  0.190 & 0.447 & 0.350  \\ 
    BASE + ParallelFuse + KEY & 43.09 & \textbf{47.70} & 49.84 & 24.92 & \textbf{28.08} & 29.82 & 39.00 & 43.08 & 45.12 & 0.206 & 0.130 & 0.337 & 0.242   \\ 
    BASE + StackFuse + KEY &43.87 & 47.71 & \textbf{49.96} & 25.50 & 28.05 & \textbf{29.94} & 39.94 & \textbf{43.27} & \textbf{45.43} & 0.242 & 0.160 & 0.392 & 0.293   \\ 
 \hline
\end{tabular}
  \caption{Multi-Headline Generation Results}
  \label{tab:headline_gen}
\end{table*}

\noindent \textbf{Baselines}. In the following experiments, we compare various variants of the proposed keyphrase-aware models we introduced in Section~\ref{sec:model} as follows:
(1) \textbf{BASE}, as shown in Figure~\ref{fig:models} (a), which is keyphrase-agnostic and only takes the news article as input.
(2) \textbf{BASE + KEY}, as shown in Figure~\ref{fig:models} (b), which takes keyphrase and article as input.
(3) \textbf{BASE + Filter}, as shown in Figure~\ref{fig:models} (c), which takes keyphrase-filtered article as input.
(4) \textbf{BASE + StackFuse}, (5) \textbf{BASE + AddFuse}, and (6) \textbf{BASE + ParallelFuse} as shown in Figure~\ref{fig:models} (d-f), which take the keyphrase-filtered article and the original article as inputs with stack-fusing, addition-fusing, and parallel-fusing mechanism, respectively. Based on \textbf{BASE + StackFuse}, \textbf{BASE + AddFuse}, and \textbf{BASE + ParallelFuse}, we further use the keyphrase as their additional inputs, like \textbf{BASE + KEY}. Then we obtain three additional variants (7) \textbf{BASE + StackFuse + KEY}, (8) \textbf{BASE + AddFuse + KEY}, and (9) \textbf{BASE + ParallelFuse + KEY}. 
In addition to \textbf{BASE}, We also compare four other keyphrase-agnostic baselines as follows. (10) \textbf{PT-NET}, the original pointer-generator network~\cite{see2017get} , which are widely used in text summarization and headline generation tasks. (11) \textbf{SEASS}~\cite{zhou2017selective}, the GRU-based~\cite{cho2014learning} sequence-to-sequence model with selective encoding mechanism, which is widely used in text summarization. (12) \textbf{Transformer + Copy}~\cite{vaswani2017attention,gu2016incorporating}, which has the same architecture hyperparameters as \textbf{BASE}, the only difference is that it does not use BERT to initialize the encoder. (13) \textbf{BASE + Diverse}, which applies diverse decoding~\cite{li2016simple} in beam search to \textbf{BASE} during inference to improve the generation diversity for multiple headlines generation. To sum up, there are a total of 13 models to compare.

\noindent \textbf{Implementation and Hyperparameters}.
The encoder and the decoder of BASE have the same architecture hyperparameters as BERT-base.
All the variants of keyphrase-aware headline generation models also have the same architecture hyperparameters as BASE.
The only difference among them is their computation steps of each block in the decoder, as shown in Figure~\ref{fig:models}.
We follow the same training strategy in~\citet{vaswani2017attention} for training.
And the implementation of them are based on Tensor2Tensor\footnote{\url{https://github.com/tensorflow/tensor2tensor}.}.
The implementations of PT-NET\footnote{\url{https://github.com/abisee/pointer-generator}.} and SEASS\footnote{\url{https://github.com/magic282/SEASS}.} are based on their open-source code.

\subsubsection{Multi-Headline Generation} \label{sec:multi_headline}
In this experiment, we only give models the news articles without the golden keyphrases.
We use SLOT to generate top-$K$ keyphrases for each article.
Then each keyphrase-aware generation model using them to generates $K$ different keyphrase-relevant headlines.
For keyphrase-agnostic baselines, we apply the beam search to generate top $k$ headlines for each article.
We also apply the diverse decoding to BASE as a strong baseline (BASE + Diverse) for further comparison.
The diversity penalty is set to be 1.0. 
It should be noted that we can also apply the diverse decoding to our keyphrase-aware models to further improve diversity.

\textbf{Metrics}. Following~\citet{li2015diversity}, we use \textit{Distinct-1} and \textit{Distinct-2} (the higher the better) to evaluate diversity, which report the degree of the diversity by calculating the number of distinct unigrams and bigrams in generated headlines for each article. 
Since randomly generated headlines are also highly diverse, we measure the quality as well.
As we discussed in Section~\ref{sec:intro}, one news article can have multiple reasonable headlines.
However, each article in our test set has only one human written headline, which may only focus on one keyphrase of the news article.
We should emphasize that there may be only one generated headline that focuses on the same keyphrase of the human-written headline, while others focus on distinct keyphrases.
It is thus not reasonable if we use the same human-written headline as the ground-truth to evaluate all generated headlines.
We assume that if the headlines generated by the model are high-quality and diverse, there would be a higher probability that one of the headlines is closer to the single ground-truth.
Therefore, we report the highest ROUGE score among the multiple generated headlines for each article.
This criterion is similar to top-$K$ errors~\cite{he2016deep} in image classification tasks.
We report the results with $K$=1, 3, 5.

\textbf{Results}.
Table~\ref{tab:headline_gen} presents the results. For diversity, we can see that all of our keyphrase-aware generation models performs significantly better than other keyphrase-agnostic baselines in both \textit{Distinct-1} and \textit{Distinct-2} metrics for all $K$.
After using the diverse decoding, BASE + Diverse achieves higher diversity.
Nevertheless, the diversity is still lower than most keyphrase-aware generation models.
As expected, BASE + Filter achieves the highest diversity, and BASE + KEY achieves the lowest diversity among the variants of our keyphrase-aware generation models.
For quality, except BASE, there is still a big gap between other keyphrase-agnostic baselines and our keyphrase-aware generation models. 
Except for BASE + Filter, all of our keyphrase-aware generation models achieve higher ROUGE scores than BASE and BASE + Diverse (see the last 6 lines in Table~\ref{tab:headline_gen}).
These results show that our keyphrase-aware generation models can effectively generate high-quality and diverse headlines.

\begin{table*}[t]
\centering
\scriptsize
  \begin{tabular}{|l|c|c|c|c|c|c|c|c|c|c|c|c|c|c|}
    \hline
    \multirow{2}{*}{Method} & \multicolumn{3}{c|}{mAP@1} & \multicolumn{3}{c|}{mAP@3} & \multicolumn{3}{c|}{mAP@5} & \multicolumn{3}{c|}{mAP@10} \\ \cline{2-13} 
    & k=1 & k=3 &k=5 & k=1 & k=3 &k=5 & k=1 & k=3 &k=5 & k=1 & k=3 & k=5 \\
    \hline
    HUMAN & 49.25 & - & - & 63.94 & - & - & 69.40 & - & - & 75.70 & - & - \\\hline
    PT-GEN      & 32.08 & 35.50 & 37.01 & 45.23 & 49.36 & 50.91 & 50.51 & 54.61 & 56.30 & 56.77 & 60.93 & 62.51  \\
    SEASS      & 28.82 & 31.66 & 32.85 & 42.09 & 45.27 & 46.60 & 47.66 & 50.59 & 52.11 & 54.51 & 57.74 & 59.13  \\
    Transformer + Copy      & 37.55 & 41.88 & 43.46 & 51.09 & 55.99 & 57.72 & 56.54 & 61.45 & 63.06 & 62.58 & 67.21 & 69.04  \\ \hline
    BASE      & 39.43 & 43.80 & 45.54 & 53.77 & 58.35 & 60.16 & 59.29 & 63.88 & 65.70 & 65.51 & 69.92 & 71.68   \\
    BASE + Diverse      & 39.30 & 45.02 & 47.07 & 53.69 & 60.05 & 62.18 & 59.10 & 65.49 & 67.61 & 65.37 & 71.65 & 73.73  \\ \hline
    BASE + Filter     & 34.30 & 43.02 & 45.91 & 48.28 & 58.07 & 61.61 & 53.91 & 63.78 & 67.42 & 60.41 & 70.51 & 74.06   \\
    BASE + KEY     & 39.38 & 45.04 & 46.81 & 53.81 & 60.05 & 61.91 & 59.39 & 65.75 & 67.59 & 65.59 & 71.99 & 73.97   \\ \hline 
    BASE + AddFuse  & 39.95 & \textbf{46.64} & 48.72 & 54.59 & \textbf{61.90} & 63.92 & 60.20 & \textbf{67.63} & \textbf{69.64} & 66.56 & \textbf{73.95} & 75.96   \\ 
    BASE + ParallelFuse & 39.82 & 46.57 & \textbf{49.03} & 54.44 & 61.80 & \textbf{64.27} & 59.91 & 67.50 & 69.85 & 66.47 & 73.82 & \textbf{76.08}     \\ 
    BASE + StackFuse & 40.11 & 45.96 & 47.95 & 54.52 & 61.05 & 63.05 & 60.11 & 66.69 & 68.70 & 66.57 & 72.91 & 74.86    \\ \hline 
    BASE + AddFuse + KEY & 39.19 & 46.01 & 48.55 & 53.73 & 61.32 & 63.70 & 59.34 & 66.99 & 69.38 & 65.70 & 73.54 & 75.77    \\ 
    BASE + ParallelFuse + KEY & \textbf{40.24} & 46.46 & 48.66 & \textbf{54.82} & 61.64 & 63.68 & \textbf{60.46} & 67.24 & 69.30 & \textbf{67.01} & 73.70 & 75.53    \\
    BASE + StackFuse + KEY & 39.78 & 46.33 & 48.55 & 54.28 & 61.43 & 63.62 & 59.95 & 67.07 & 69.21 & 66.20 & 73.35 & 75.37    \\ 
 \hline
\end{tabular}
  \caption{News Article Retrieval Results}
  \label{tab:retrieval}
\end{table*}

\subsubsection{News Article Retrieval} \label{sec:retrieval}
To further evaluate the quality and diversity of the generation, we design an experiment that uses a search engine to help us verify the diversity and quality of the generated headlines.
It should be noted that the main purpose of this experiment is not to improve the performance of the search engine, but to measure the quality and diversity of the generated multiple headlines through a real-world search engine.
We first collect the data pairs of the news article and its related user search query $\langle X, Q \rangle$ in the following way.
If the article $X$ is returned by the search engine based on a user query $Q$ and the user clicks on the article $X$, then we take the query and the article as a data pair $\langle X, Q \rangle$.
After collection, each article $X$ in the test set has 10 different related user queries on average.
The article $X$ is used as the ground-truth for $Q$ in the following evaluation.
We replace the search key in the original search engine for each article with the $K$ generated multi-headlines.
Also, we re-build the indexes of the search engine that contains 10,000 news articles in the test set.
Then we re-use the user search queries to retrieve the article.
We believe that if the generated multi-headlines have high diversity and quality, then given different user queries, there should be a high probability that the golden article can be retrieved.

\textbf{Metrics}.
We use the mean average precision (mAP), which is widely used for information retrieval as a metric.
We report the results of mAP@N ($N$=1, 3, 5, and 10) which test the average probability that the golden article is ranked by the search engine to the top $N$.
Using the human-written headline (HUMAN) as the search key is evaluated as a strong baseline.
We also compare the performance of using a different number of headlines ($K$=1, 3, and 5).
It should be noted that increasing the number of headlines as the search key does not ensure the improvement of the mAP, because the number of search keys of all other articles will also be increased.
If the generated multi-headlines are not good enough, it will introduce noise and even cause mAP decreasing.

\textbf{Results}. Table~\ref{tab:retrieval} presents the results.
Similarly, our models perform much better than other keyphrase-agnostic baselines.
In most cases, BASE + Diverse outperforms BASE, but still performs worse than 7 keyphrase-aware generation models (see the last 7 lines in Table~\ref{tab:retrieval}).
Generally, with the number of headline $K$ increases, we can see that the performance of our keyphrase-aware generation models improves much higher than other baselines.
We find that the mAP@10 of BASE + ParallelFuse ($K$=5) achieves 76.08, which is even better than HUMAN.
These results demonstrate that our keyphrase-aware generation models can generate high-quality and diversity headlines.

\begin{figure*}[t] 
  \centering
  \includegraphics[scale=0.4]{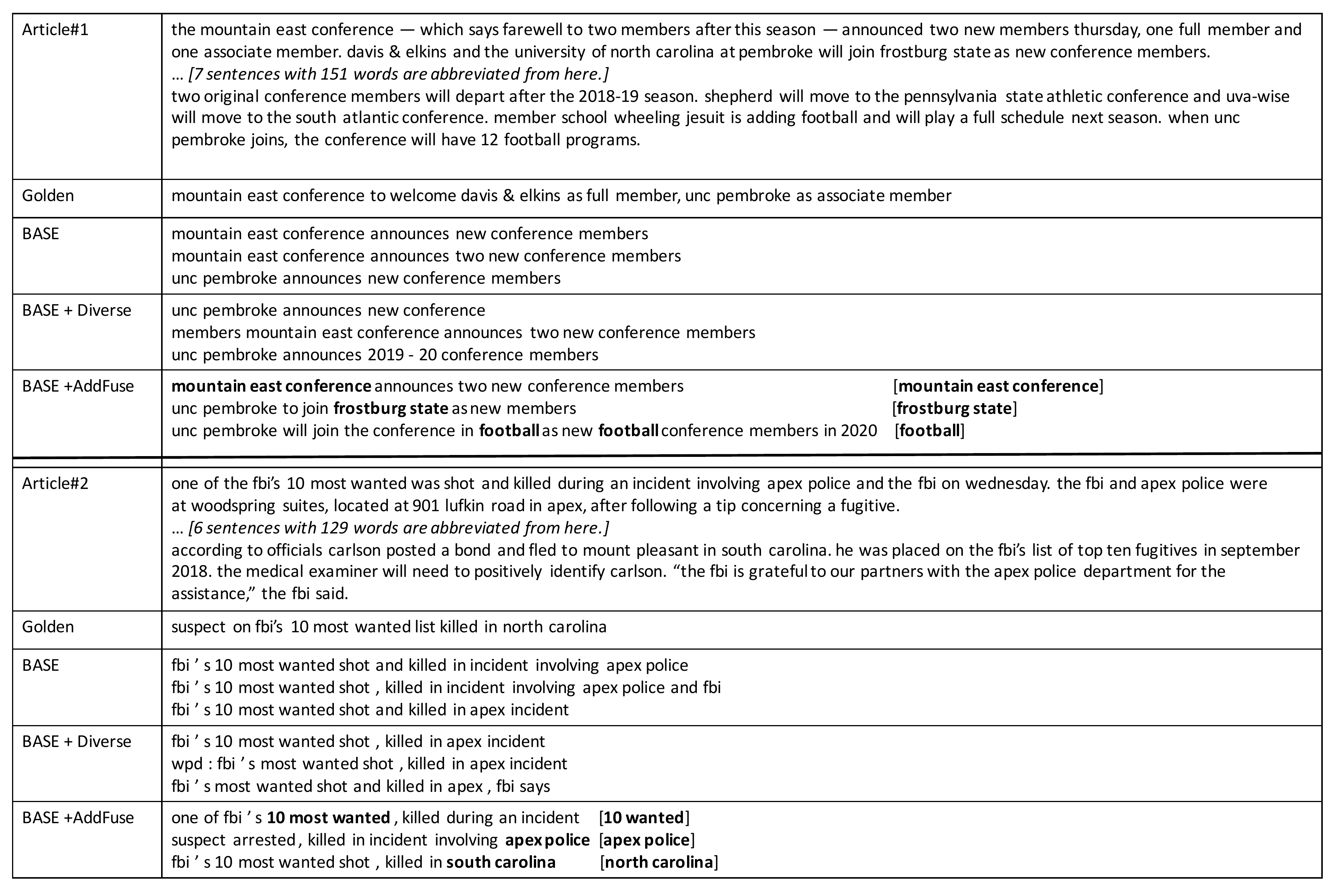}
  \caption{Examples of original articles, golden headlines and multiple generated outputs by BASE, BASE + Diverse and BASE + AddFuse. Each generated keyphrase is shown at the end of each generated headline.} \label{fig:examples}
\end{figure*}

\subsubsection{Human Evaluation} 
At a similar level of diversity, we want to investigate the quality of the headlines generated by our keyphrase-aware headline generation model compared to BASE + Diverse. Since the Distinct-1 and Distinct-2 of BASE + Diverse and BASE + StackFuse are close, we compare the quality of them through human evaluation. We randomly sample 100 articles from the test set, let each model generate 3 different headlines. We also mix a random headline and the golden headline for each article, and thus each article has 8 headlines.
Three experts are asked to judge whether each headline could be used as the headline of the news.
If more than two experts believe that it can be used as the headline of the news, then this headline is considered qualified.
The results of the qualified rate of golden, BASE + Stack, BASE + Diverse, and random are 91.8\%, 62.6\%, 36.2\%, and 0.0\%, respectively. 
These results show that the quality of BASE + StackFuse is also higher than BASE + Diverse. We present some examples for comparison as shown in Figure~\ref{fig:examples}.

\section{Related Works}
News headline generation is a subtask of summarization which has been extensively studied recently~\cite{rush2015neural,takase2016neural,ayana2016neural,tan2017neural,zhou2017selective,higurashi2018extractive,zhang2018question,murao2019case}. 
With the rapid development of neural networks, various neural models have been successfully used in the headline generation task.
\citet{rush2015neural} propose an attention-based neural network for headline generation. 
\citet{takase2016neural} propose a method based on encoder-decoder architecture and design an AMR encoder for headline generation.  
To take evaluation metrics into consideration,~\citet{ayana2016neural} apply minimum risk training method to the generation model. \citet{tan2017neural} propose a coarse-to-fine method, which first extracts the salient sentences and then generates the headline based on these sentences.
\citet{zhou2017selective} propose a method which divides the process of headline generation into three phases: a sentence encoder, a selective gate network for sentence selection, and a headline decoder.
\citet{higurashi2018extractive} propose an extractive headline generation method, different from previous works that target the headline generation for the articles, this work focus on the task of headline generation for the community question answering forums. Due to lack of supervised training data, they propose a learning-to-rank based method to extract the essential substring from a question and use this substring as the headline of the forums.
\citet{zhang2018question} propose a method for question headline generation, which designs a dual-attention seq2seq model.

However, most previous headline generation methods focus on one-to-one mapping, and the headline generation process is not controllable. In this work, we focus on the news multi-headline generation problem and design a keyphrase-aware headline generation method. Different information aware methods have been successfully used in natural language generation tasks~\cite{zhou2017mechanism,zhou2018commonsense,wang2017group}, such as responses generation in the dialogue system.  Similar to our task, responses generation in a dialogue system is also a one-to-many problem,
\citet{zhou2017mechanism} propose a mechanism-aware seq2seq model for controllable response generation. They model different mechanisms as latent embeddings and learn the latent embeddings in their seq2seq model. Incorporating these mechanisms, their model can generate controllable responses.
\citet{zhou2018commonsense} propose a commonsense knowledge aware conversation generation method. More concretely, in the first stage, their model retrieves subgraphs from a knowledge base, and the model encodes the subgraphs using a  dynamic graph neural network to facilitate better conversation generation in the second stage.
\citet{wang2017group} propose an encoder-decoder based neural network for response generation. To our best knowledge, we are the first to consider keyphrase-aware mechanism on news headline generation and build the first keyphrase-aware news headline corpus. 

\section{Conclusion}
In this paper, we demonstrate how to enable news headline generation systems to be aware of keyphrases such that the model can generate diverse news headlines in a controlled manner.
We also build a first large-scale keyphrase-aware news headline corpus, which is based on mining the keyphrases of users' interests in news articles with user queries.
Moreover, we propose a keyphrase-aware news multi-headline generation model that contains a multi-source Transformer decoder with three variants of attention-based fusing mechanisms. 
Extensive experiments on the real-world dataset show that our approach can generate high-quality, keyphrase-relevant, and diverse news headlines, which outperforms many strong baselines.

\section*{Acknowledgment}
This work is supported in part by the National Natural Science Fund for Distinguished Young Scholar under Grant 61625204, and in part by the State Key Program of the National Science Foundation of China under Grant 61836006. 

\bibliographystyle{acl_natbib}
\bibliography{emnlp2020}

\end{document}